# Deep Blind Compressed Sensing

Shikha Singh[1], Vanika Singhal[2] and Angshul Majumdar[3]

[1]*IIITD*
*Okhla Phase 3*
*New Delhi 110020, India*
*shikhas@iiitd.ac.in*

[2]*IIITD*
*Okhla Phase 3*
*New Delhi 110020, India*
*vanikas@iiitd.ac.in*

[3]*IIITD*
*Okhla Phase 3*
*New Delhi 110020, India*
*angshul@iiitd.ac.in*

*Abstract*: This work addresses the problem of extracting deeply learned features directly from compressive measurements. There has been no work in this area; existing deep learning tools only give good results when applied on the full signal (that too usually after pre-processing). These techniques require the signal to be reconstructed first. In this work we show that by learning directly from the compressed domain, considerably better results can be obtained. This work extends the recently proposed framework of deep matrix factorization in combination with blind compressed sensing; hence the term 'deep blind compressed sensing'. Simulation experiments have been carried out on imaging via single pixel camera, under-sampled biomedical signals (arising in wireless body area network) and compressive hyperspectral imaging. In all cases, the superiority of our proposed deep blind compressed sensing can be envisaged.

## 1. Introduction

Compressed Sensing (CS) encompasses the theory, algorithms and applications of solving an under-determined system of linear equations when the solution is known to be sparse. Over the past decade CS has found many applications in image and signal processing. Especially when the signal / image is acquired in some other (measurement) domain, e.g. MRI in Fourier domain, X-Ray CT in Radon, etc.; and the task is to recover the signal from the measurements.

CS assumes that the sparsifying transform is known (wavelet, DCT etc.). Such mathematical transforms are versatile; they can sparsely represent a large class of signals in a sparse fashion. However, practitioners and engineers are not interested in there generalizability; they are more interested in solving their problem in the best possible way. The reconstruction ability of CS is guided by the sparsity of the signal in the transform – the sparser (larger the number of zeroes) it is, the better is the reconstruction. It has been found that better results can be achieved when the sparsifying operator is learnt (from data) instead of being designed (e.g. wavelet, DTC etc.).

Dictionary learning concerns itself about learning such sparsifying transforms from data. It has been empirically seen that inverse problems, solved via dictionary learning, yields significantly better results than CS (with designed sparsifying transforms). Blind Compressed Sensing (BCS) combines CS with dictionary learning. It learns the sparsifying basis while reconstructing the signal.

Early studies in dictionary learning were not interested in solving inverse problems but in computer vision. Even today more than half of the papers published in this area are on vision related research. It is used for feature extraction in vision problems. Basic dictionary learning is a matrix factorization problem. The training data is expressed as a product of the dictionary matrix and the coefficient matrix. Dictionary learning has been expressed in terms of latent factor model, where the coefficient / features are the latent factors corresponding to the input / observed samples. Apart from dictionary learning / matrix factorization, there are several other approaches to latent factor modelling. Restricted Botzmann Machine (RBM) and Autoencoder (AE) fall in the same category. Today, the term 'representation learning' is used (in a broader sense) instead of 'latent factor model'.

RBM and AE are routinely used for training deep neural networks. Stacking RBMs lead to Deep Belief Network and Deep Boltzmann Machine and nesting AE, one inside the other leads to stacked autoencoders. For all cases, the representation from the final / deepest layer is used to train some classifier (soft-max or logistic regression). A recent study introduced the concept of deep dictionary learning (DDL). Here the first layer of dictionary and coefficients are learnt from the input data; the subsequent layers learn from the coefficients of the previous layer as inputs - this is a typical greedy learning paradigm. DDL can accommodate non-linear activation functions. A recent work on deep matrix factorization (DMF) has proposed simultaneous learning of multiple layers of matrices. It turns out to be a special case of DDL where the activation functions are linear identity.

All prior deep learning tools are used on the complete data, i.e. images or signals in the physical domain. In this work, we address the problem of learning from the compressed measurement domain. Such problems may arise in scientific problems, e.g. images captured by single pixel camera, or analysis of compressed biomedical signals in wireless body area networks. One can follow the usual piecemeal approach of reconstruction followed by application of standard deep learning tools for analysis. In this work, we will show that the deep matrix factorization can be extended to extract features directly from the compressed domain and the analysis results are considerably better than piecemeal approach.

## 2. Brief Literature Review

In the first sub-section we will briefly discuss about compressed sensing, dictionary learning and blind compressed sensing. In the second sub-section we will discuss about deep learning.

### 2.1. Compressed Sensing, Dictionary Learning and Blind Compressed Sensing

Compressed Sensing (CS) is concerned about solving an under-determined system of linear equations when the solution is known to be sparse.

$$y_{m \times 1} = A_{m \times n} x_{n \times 1}, \ m < n \tag{1}$$

In general such a system has infinitely many solutions. But in most cases, a sparse solution is unique [1]. The solution can be recovered by an $l_1$-minimization problem.

$$\min_{x} \|x\|_1 \text{ subject to } y = Ax \tag{2}$$

Real signals are hardly ever sparse; but most of them have a compressible representation in some domain (say $S^1$). This allows expressing a real-life compressed sensing problem.

$$\min_{\alpha} \|\alpha\|_1 \text{ subject to } y = AS^T\alpha \tag{3}$$

In most cases, the signal is not exactly sparse but is compressible. In such a case, one solves the following problem.

$$\min_{\alpha} \|y - AS^T\alpha\|_2^2 + \lambda\|\alpha\|_1 \tag{4}$$

This is a relaxation from the equality constraint to account for compressibility and lossy reconstruction.

So far, it has been assumed that the sparsifying transform is given. It is usually a designed transform such as wavelet, curvelet, DCT etc. In CS, the quality of reconstruction is dependent on the sparsity of the solution[2]; this in turn depends on the sparsifying transform. It has been observed that learning a sparsifying transform adaptively from data yields significantly better results than the designed ones. This leads to the topic of dictionary learning (DL) [2, 3].

DL is a synthesis formulation where a basis / dictionary is learnt such that it can synthesize / regenerate the data from the learnt coefficients. This is expressed as,

$$X = DZ \tag{5}$$

where X is the training data, D the learnt dictionary and Z the coefficients / features / latent representation. In CS, one is interested in sparse coefficients, therefore the learning is framed as,

$$\min_{D,Z} \|X - DZ\|_F^2 + \lambda\|Z\|_1 \text{ such that } \|D_i\|_2^2 = 1 \tag{6}$$

We have abused the notation slightly, the $l_1$-norm is defined on the vectorized version of Z.

The additional constraint on the unit normed columns of the dictionary ensures that the solution is not degenerate (very large D and very small Z). Problem (6) is solved using alternating minimization; the dictionary is updated assuming the coefficients to be constant and the features are updated assuming the dictionary to be constant.

A simple way to combine dictionary learning with CS is to learn the sparsifying basis / dictionary from a training data and apply it as a sparsifying transform for CS reconstruction. However, such a piecemeal technique would not guarantee sparse representation of the test (to be reconstructed) signal in a dictionary learnt on some other training data. Therefore in practice the dictionary is learnt from the data itself. The formulation is expressed as,

---

[1] Analysis : $\alpha = Sx$

Synthesis : $x = S^T\alpha$

[2] It also has dependency on the incoherence between the sparsifying and measurement basis. But later studies found the dependency to be much less pronounced than initially assumed to be.

$$\min_{D,Z,x} \|y - Ax\|_2^2 + \sum_i \|P_i x - D z_i\|_2^2 + \lambda \|z_i\|_1 \text{ such that } \|D_i\|_2^2 = 1 \tag{7}$$

Here $P_i$ is the patch extraction operator. Therefore term $\sum_i \|P_i x - D z_i\|_2^2 + \lambda \|z_i\|_1$ such that $\|D_i\|_2^2 = 1$ is the standard dictionary learning operation. The first term $\|y - Ax\|_2^2$ corresponds to data consistency. This (7) is a generic formulation. For denoising problems [4] $A$=identity and for reconstruction [5] it is the measurement operator.

DL is applicable to single measurement vector problems since it is a patchwise operation. Blind compressed sensing (BCS) [6] is applicable for multiple vector problems. It learns a basis / dictionary (D) and coefficients (Z) from the compressed measurements (Y).

$$Y = AX = ADZ \tag{8}$$

Here $X = [x_1 | ... | x_N]$ and $Y = [y_1 | ... | y_N]$; this is a multiple measurement vector recovery problem. The task is to reconstruct X given the compressive measurements Y. Instead of assuming sparsity of X in some basis, BCS learns it on the fly. This is expressed as,

$$\min_{D,Z} \|Y - ADZ\|_F^2 + \lambda \|Z\|_1 + \mu \|D\|_F^2 \tag{9}$$

Here the $l_1$-norm on Z and the Frobenius-norm on D prevents the degenerate solution. Once the dictionary and the coefficients are learnt, the signal is recovered by $X=DZ$.

## 2.2. Deep Learning

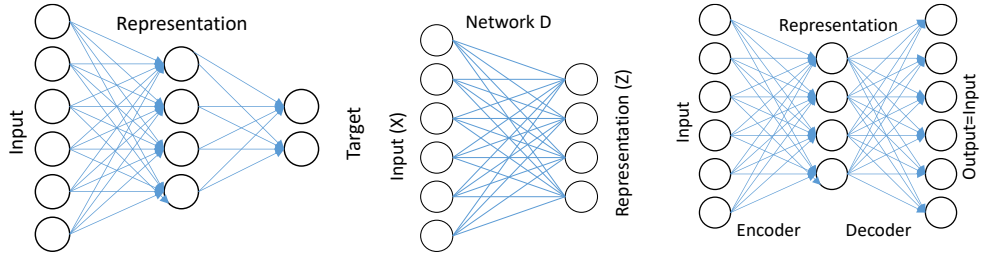

Figure 1. Left – Single Layer Neural Network; Mid – RBM; Right - Autoencoder

Figure 1 (left) shows the diagram of a simple neural network with one representation (hidden) layer. The problem is to learn the network weights between the input and the representation and between the representation and the target. This can be thought of as a segregated problem – i) learning from input to representation and ii) learning from representation to output. Learning the mapping between the representation and the target is straightforward. The challenge is to learn the network weights (from input) and the representation. This is the study of representation learning.

Restricted Boltzmann Machine (RBM) [7] is one technique to learn the representation layer. It learns by maximizing the Botlzmann cost function.

$$p(D, Z) = e^{Z^T D X} \tag{10}$$

Broadly speaking, RBM learning is based upon maximizing the similarity between the projection of the data and the representation, subject to the usual constraints of probability.

The first generation of RBMs could not handle inputs other than binary. Later this problem was partially addressed by Gaussian-Bernoulli RBM [8] that could handle inputs ranging between 0 and 1. Multiple layers of RBM are stacked (loosely speaking) to form a DBN [9]. Once the deep architecture is learnt, the representation / features from the final layer can be used to learn a soft-max classifier, thus completing the deep neural network or can be used to train other classifiers.

The other prevalent technique to train the representation layer of a neural network is by autoencoder [10]. The architecture is shown in Figure 1 (right).

$$\min_{W,W'} \|X - W'\phi(WX)\|_F^2 \tag{11}$$

The cost function for the autoencoder is expressed above. W is the encoder, and W' is the decoder. The autoencoder learns the encoder and decoder weights such that the reconstruction error is minimized. Essentially it learns the weights such that the representation $\phi(WX)$ retains all the information of the data needed for its reconstruction. Once the autoencoder is learnt, the decoder portion of the autoencoder is removed and the target is attached after the representation layer. Multiple units of such autoencoders are nested inside the other to form SAE [11].

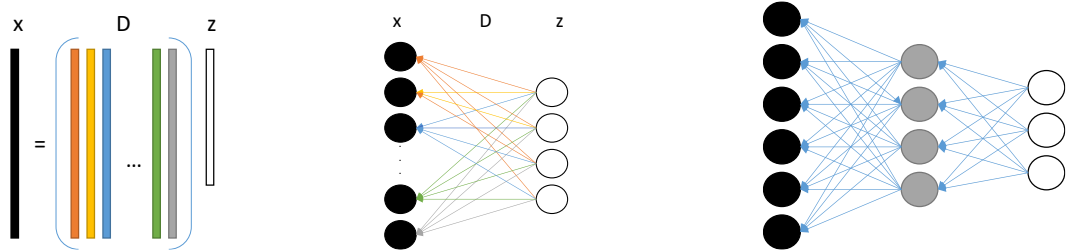

Figure 2. Left – Dictionary Learning; Mid – Neural Network Interpretation; Left – Deep Dictionary Learning

The interpretation for dictionary learning is different. It learns a basis (D) for representing (Z) the data (X). The columns of D are called 'atoms'. In this work, we look at dictionary learning in a different manner. Instead of interpreting the columns as atoms, we can think of them as connections between the input and the representation layer. To showcase the similarity, we have kept the color scheme intact in Figure 2 (mid). Unlike a neural network which is directed from the input to the representation, the dictionary learning kind of network points in the other direction – from representation to the input.

Based on the neural network type interpretation of dictionary learning [12] proposed deeper architectures using dictionary learning as the basic building block. Expressing the data via multiple levels of dictionaries,

$$X = D_1\varphi\big(D_2\varphi(\ldots\varphi(D_N Z))\big) \tag{12}$$

Here φ are the activation functions. The learning is expressed as,

$$\min_{D_1,\ldots D_N, Z} \|X - D_1\varphi\big(D_2\varphi(\ldots\varphi(D_N Z))\big)\|_F^2 + \mu\|Z\|_1 \tag{13}$$

In [12] a greedy layer by layer approach is followed for solving (13). Computations become easy; since shallow (single layer) dictionary learning is a well researched area. But the problem with this approach is that there is no feedback from deeper to shallower layers.

Deep dictionary learning is a generalization of deep matrix factorization [13] with non-linear activations.

## 3. Proposed Formulation

In this work we propose to directly learn deep features from the compressed domain. Let us assume that Y is the compressed representation of the training data X. This is expressed as,

$$Y = AX, \ Y = [y_1 | ... | y_N] \text{ and } X = [x_1 | ... | x_N] \tag{13}$$

In BCS a single level of dictionary is learnt to represent the data (X=DZ). Following deep matrix factorization, we propose to learn multiple levels of dictionaries to express the data, i.e.

$$X = D_1 D_2 ... D_M Z \tag{14}$$

It must be noted that, even though there is no non-linear activation function, (14) is not the same as learning a shallow dictionary learning model with $D = D_1 D_2 ... D_M$. This is because dictionary learning is a bi-linear problem and deep dictionary learning is a multi-linear problem and it is not possible to collapse all the layers of dictionary into a single one. For more details please refer [12, 13].

Instead of learning from the original data, we propose learning from the compressed domain. This leads to,

$$Y = AX = AD_1 D_2 ... D_M Z \tag{15}$$

The learning problem is expressed as,

$$\min_{D_1, D_2, ..., D_M, Z} \|Y - AD_1 D_2 ... D_M Z\|_F^2 + \lambda \|Z\|_1 \tag{16}$$

This is a multi-linear problem and can be solved using alternating directions (the same has been used in [13]). The sub-problems are,

$$P1: \min_{D_1} \|Y - AD_1 D_2 ... D_M Z\|_F^2$$

$$P2: \min_{D_2} \|Y - AD_1 D_2 ... D_M Z\|_F^2$$

.
.
.

$$PM: \min_{D_M} \|Y - AD_1 D_2 ... D_M Z\|_F^2$$

$$PM+1: \min_{Z} \|Y - AD_1 D_2 ... D_M Z\|_F^2 + \lambda \|Z\|_1$$

Each of the sub-problems are linear. Sub-problems P1 to PM (for solving the multiple levels of dictionaries) are simple least square minimization problems and can be solved using conjugate gradient. Sub-problem PM+1, for solving the coefficients, can be solved using the iterative soft thresholding algorithm.

After every iteration, the columns of the dictionaries are normalized. This prevents the degenerate solution where the dictionaries are too large and the coefficients too small. The other extreme (small Z and large $D_i$'s) have been accounted for in the formulation by the penalty on Z.

The problem (16) could have been solved using multiplicative updates as well. But alternating minimization is a simpler approach.

## 4. Experimental Results

### 4.1. Single Pixel Imaging

Our proposed technique is unsupervised. Therefore it is fair to compare with unsupervised deep learning techniques only; hence we have considered – stacked denoising autoencoder (SDAE) [14], sparse stacked autoencoder (SSAE) [15], contractive autoencoder (CAE) [16], deep belief network (DBN) [9] and sparse DBN (SDBN) [17]. Our proposed method is applied on the compressed samples. But the rest are applied on the data under two scenarios. In the first one, they are directly applied on the compressed measurements and in the second one, the data is reconstructed using compressed sensing and then representation learning is carried out. After CS reconstruction the SSIM (between reconstructed and original) is 0.95 on average.

Here our goal is to test the representation learning capacity of different techniques. Therefore we use a simple nearest neighbour classifier.

We carry out evaluation on three different problems. The first one is on image recognition from measurements captured by single pixel camera [18]. Three benchmark deep learning datasets are used for evaluation – MNIST, CIFAR-10 and SVHN (Street View House Number). The number of sparse binary projections is 25% of the size of the image.

Table 1. Results from Single Pixel Imaging

| Dataset | Proposed | Learning from Compressed Domain | | | | | Learning after CS Reconstruction | | | | |
|---|---|---|---|---|---|---|---|---|---|---|---|
| | | SDAE | SSAE | CAE | DBN | SDBN | SDAE | SSAE | CAE | DBN | SDBN |
| MNIST | **98.96** | 96.33 | 95.56 | 96.49 | 96.05 | 96.94 | 97.05 | 95.94 | 96.58 | 97.44 | 97.33 |
| CIFAR-10 | **85.06** | 76.62 | 78.29 | 77.51 | 76.96 | 82.60 | 77.90 | 79.55 | 78.27 | 75.30 | 83.02 |
| SVHN | **94.55** | 90.11 | 90.05 | 90.72 | 87.29 | 91.00 | 91.60 | 90.42 | 92.13 | 88.70 | 91.45 |

Our method clearly outperform others by a significant margin – be it learning directly in the compressed domain or learning after reconstruction. For others, feature extraction followed by reconstruction yields slightly better results than direct feature extraction in the compressed domain. Please note that the results obtained here cannot be compared with the ones in prior studies, this is because the learning here is from a compressed / reconstructed samples and not the original ones as done in literature.

### 4.2. Compressive ECG Acquisition for WBAN

In the next set of experiments, we show results on the problem of biomedical signal compression and transmission over wireless body area network [19]. In such a scenario, the signals (EEG, ECG, MEG etc.) are randomly under-sampled (not projected onto a CS type projection matrix) – this reduces energy consumption for both signal acquisition and compression at the sensor nodes. The under-sampled signal is transmitted to a base station, where BCS is used for recovering the signals for further analysis.

In this study we carry out simulation experiments on transmission and classification (five types of beat classes of arrhythmia as recommended by Association for Advancement of Medical Instrumentation) of ECG signals. The classes are non-ectopic beats (N), supra-ventricular ectopic beats (S), ventricular ectopic beats (V), Fusion beats (F) and unclassifiable beats (Q). The experiments are carried out on the most popular test dataset – the MIT-BIH Arrhythmia dataset from www.physionet.org. Owing to the relative sparsity of samples in the last two classes (F and Q), we follow the more recent AAMI2 protocol [20] where the last two classes are merged with V.

To the best of our knowledge there is only a single comprehensive study on deep learning based arrhythmia classification [21]. Standard features are extracted (for details see [21]), followed by deep learning using SDAE and DBN. Both SDAE and DBN has fine-tuned soft-max classifier. In this work, we under-sample the ECG signals by 25%. Our proposed method directly learns on the compressed samples, there is no need for reconstruction and feature extraction. We train an SDAE and DBN for feature extraction and classification from the under-sampled domain. We also reconstruct the signal using standard BCS [19] followed by feature extraction and deep learning as in [21]. The results are shown in the following table. The specificity and sensitivity for each class are reported and the overall accuracy is reported. This is the standard protocol. For our method a support vector machine with rbf kernel is used.

Table 2. ECG Arrhythmia Classification

| Method | F | | S | | V | | Accuracy |
|---|---|---|---|---|---|---|---|
| | Spec. | Sens. | Spec. | Sens. | Spec. | Sens. | |
| Proposed | **100** | **67.2** | 16.9 | **100** | **90.1** | **100** | 97.0 |
| SDAE on under-sampled signals | 87.3 | 33.9 | 0 | 82.6 | 39.8 | 81.6 | 85.6 |
| DBN on under-sampled signals | 60.2 | 25.9 | 0 | 51.3 | 16.8 | 28.1 | 58.7 |
| SDAE on reconstruction after feature extraction | 96.9 | 62.8 | 20.8 | **100** | 83.2 | 100 | 93.8 |
| DBN on reconstruction after feature extraction | 95.7 | 66.5 | **23.9** | **100** | 86.9 | **100** | 94.8 |

We find that our proposed method yields the best results even though it operates in the most challenging circumstance – no reconstruction or feature extraction. SDAE and DBN in this scenario do significantly worse. Prior study [21] only show results comparable to ours (still trailing by a considerable margin) when the signals are reconstructed and hand-crafted features are obtained. These results cannot be compared with [21] since we work with under-sampling and reconstruction and not the fully sampled data.

### 4.3. Compressive Hyperspectral Imaging

In the final set of experiments, we show hyper-spectral image classification results. There are three recent papers on this topic [22-24]; all of them are straightforward applications of SDAE [22], DBN [23] and CNN [24]. We compare our proposed technique on the standard evaluation datasets – Indian Pines which has 200 spectral reflectance bands after removing bands covering the region of water absorption and 145*145 pixels of sixteen categories, and the Pavia University scene which has 103 bands of 340*610 pixels of nine categories.

Compressive hyper-spectral image acquisition is simulated [25]. As we have done before, deep learning is carried out in the compressed domain and in the pixel domain after compressed sensing reconstruction. The average SSIM after reconstruction from 25% samples is 0.87. After reconstruction, [22, 23] uses spatio-spectral features followed by deep learning. It is not possible to employ CNN in the compressed domain. The results are shown in table 3.

Prior studies on deep learning based classification assumed an overtly optimistic scenario [22 – 24] – they assumed 80% for training and 20% for testing. In this work we follow the more standard (albeit challenging) evaluation protocol on these datasets. For the first dataset (Indian Pines), we randomly select 10% of the labelled data as training set and rest as testing set; for the second dataset (Pavia University) 2% of the labelled data is used for training and the rest for testing. The performance is measured in terms of the three standard measures – overall accuracy (OA), average accuracy (AA) and kappa coefficient.

Table 3. Hyper-spectral Image Classification

| Dataset | Metric | Proposed | Learning from Compressed Domain | | Learning from CS Reconstruction | | |
|---|---|---|---|---|---|---|---|
| | | | SDAE | DBN | SDAE | DBN | CNN |
| Pavia | AA | **94.09** | 81.03 | 74.88 | 85.02 | 78.50 | 87.12 |
| | OA | **96.81** | 87.89 | 78.06 | 88.26 | 86.09 | 95.34 |
| | Kappa | **0.95** | 0.88 | 0.80 | 0.90 | 0.84 | 0.94 |
| Indian Pines | AA | **73.40** | 67.78 | 65.69 | 78.33 | 73.33 | 83.19 |
| | OA | **81.93** | 70.23 | 67.38 | 86.10 | 81.79 | 90.21 |
| | Kappa | **0.82** | 0.71 | 0.66 | 0.73 | 0.67 | 0.78 |

The results corroborate the previous findings. In the compressed domain, the standard deep learning tools yield poor results. But once the signal is reconstructed and spatio-spectral features (from patch cubes of hyperspectral image followed by PCA) improve the results. However none of the prior techniques can surpass ours in any of the performance measures. CNN with its hand-tuned filters comes close, but still falls short of deep blind compressed sensing.

## 5. Conclusion

This work proposes a new deep learning tool for learning from the compressed domain. It combines blind compressed sensing with the recently proposed framework of deep matrix factorization. In blind compressed sensing, the signal is learnt from compressed measurements by learning a factorization into two matrices – one dictionary and one coefficient. In deep matrix factorization, there are multiple levels of dictionaries and one final level of coefficient matrix. In our proposed deep blind compressed sensing framework, we learn all the levels of dictionaries and the coefficient matrix from compressive measurements.

We have carried out experiments on three types of problems – imaging via single pixel camera, ECG signal analysis from sub-sampled measurements and compressive hyper-spectral imaging. For all the experiments we applied our proposed technique on the compressive measurements and compared existing deep learning tools in two different scenarios. In the first one they are applied on the compressed measurements and in the second one they are reconstructed via (CS or BCS) and deep learning is carried out after. The existing deep learning tools perform poorly in the compressed domain; after reconstruction their results improve. But in neither of the three types of experiments, can the existing deep learning tools (compared against) beat our proposed approach.

## References

[1] D. L. Donoho, "For most large underdetermined systems of linear equations the minimal l1-norm solution is also the sparsest solution", Communications of Pure and Applied Mathematics. Vol. 59 (6), pp. 797-829, 2006.

[2] B. Olshausen and D. Field, "Sparse coding with an overcomplete basis set: a strategy employed by V1?", Vision Research, Vol. 37 (23), pp. 3311-3325, 1997.

[3] D. D. Lee and H. S. Seung, "Learning the parts of objects by non-negative matrix factorization", Nature 401 (6755), pp. 788-791, 1999.

[4] M. Elad and M. Aharon, "Image Denoising Via Sparse and Redundant Representations Over Learned Dictionaries," IEEE Transactions on Image Processing, Vol.15 (12), pp. 3736-3745, 2006.

[5] J. Caballero, A. N. Price, D. Rueckert and J. V. Hajnal, "Dictionary Learning and Time Sparsity for Dynamic MR Data Reconstruction," IEEE Transactions on Medical Imaging, Vol. 33 (4), pp. 979-994, 2014.

[6] S. Gleichman, Y. C. Eldar. "Blind Compressed Sensing". IEEE Trans. on Information Theory, Vol. 57, pp. 6958-6975, 2011.

[7] R. Salakhutdinov, A. Mnih, G. Hinton, "Restricted Boltzmann machines for collaborative filtering", ICML, 2007.

[8] K.H. Cho, T. Raiko and A. Ilin, "Gaussian-Bernoulli deep Boltzmann machine," IEEE IJCNN, 2013.

[9] R. Salakhutdinov and G. Hinton, "Deep Boltzmann Machines", AISTATS, 2009.

[10] N. Japkowicz, S. J. Hanson and M. A. Gluck, "Nonlinear autoassociation is not equivalent to PCA", Neural Computation, Vol. 12 (3), pp. 531-545, 2000.

[11] Y. Bengio, "Learning deep architectures for AI", Foundations and Trends in Machine Learning, Vol. 1(2), pp. 1-127, 2009.

[12] S. Tariyal, A. Majumdar, R. Singh and M. Vatsa, "Deep Dictionary Learning", IEEE ACCESS.

[13] George Trigeorgis, Konstantinos Bousmalis, Stefanos Zafeiriou, Bjoern W.Schuller, "A Deep Semi-NMF Model for Learning Hidden Representations", ICML 2014.

[14] P. Vincent, H. Larochelle, I. Lajoie, Y. Bengio and P. A. Manzagol, "Stacked denoising autoencoders: Learning useful representations in a deep network with a local denoising criterion", The Journal of Machine Learning Research, Vol. 11, pp. 3371-3408, 2010.

[15] K. Cho, "Simple sparsification improves sparse denoising autoencoders in denoising highly noisy images", ICML, 2013.

[16] S. Rifai, P. Vincent, X. Muller, X. Glorot and Y. Bengio, "Contractive auto-encoders: Explicit invariance during feature extraction", ICML, 2011.

[17] Y. L. Boureau and Y. L. Cun, "Sparse feature learning for deep belief networks", NIPS, 2008.

[18] http://dsp.rice.edu/cscamera

[19] A. Majumdar and R. K. Ward, "Energy Efficient EEG Sensing and Transmission for Wireless Body Area Networks: A Blind Compressed Sensing Approach", Biomedical Signal Processing and Control, Vol. 20, pp. 1-9, 2015.

[20] E. José da S. Luz, T. M. Nunes, , V. H. C. de Albuquerque, J. P. Papa and D. Menotti, "ECG arrhythmia classification based on optimum-path forest", Expert Systems with Applications, Vol. 40 (9), pp. 3561-3573, 2013.

[21] M.M. Al Rahhal, Y. Bazi, H. AlHichri, N. Alajlan, F. Melgani and R.R. Yager, "Deep learning approach for active classification of electrocardiogram signals", Information Sciences, Vol. 345 (1), pp. 340–354, 2016.

[22] Y. Chen, Z. Lin, X. Zhao, G. Wang and Y. Gu, "Deep Learning-Based Classification of Hyperspectral Data", IEEE Journal of Selected Topics in Applied Earth Observations and Remote Sensing, Vol. 7 (6), pp. 2094 - 2107, 2014.

[23] Y. Chen, X. Zhao and X. Gia, "Spectral-Spatial Classification of Hyperspectral Data Based on Deep Belief Network", IEEE Journal of Selected Topics in Applied Earth Observations and Remote Sensing, Vol. 8 (6), pp. 2381 - 2392, 2015.

[24] A. Romero, C. Gatta and G. Camps-Valls, "Unsupervised Deep Feature Extraction for Remote Sensing Image Classification", IEEE Transactions on Geoscience and Remote Sensing, Vol. 54 (3), pp. 1349 - 1352, 2016.

[25] X. Lin, Y. Liu, J. Wu and Q. Dai, "Spatial-spectral encoded compressive hyperspectral imaging". ACM Transactions on Graphics, Vol. 33(6), pp. 233, 2014.